# Improving Bi-LSTM Performance for Indonesian Sentiment Analysis Using Paragraph Vector

Ayu Purwarianti
School of Electrical Engineering and Informatics
Institut Teknologi Bandung
Bandung, Indonesia
ayu@stei.itb.ac.id

Ida Ayu Putu Ari Crisdayanti
School of Electrical Engineering and Informatics
Institut Teknologi Bandung
Bandung, Indonesia
dayu.crish@gmail.com

*Abstract*—Bidirectional Long Short-Term Memory Network (Bi-LSTM) has shown promising performance in sentiment classification task. It processes inputs as sequence of information. Due to this behavior, sentiment predictions by Bi-LSTM were influenced by words sequence and the first or last phrases of the texts tend to have stronger features than other phrases. Meanwhile, in the problem scope of Indonesian sentiment analysis, phrases that express the sentiment of a document might not appear in the first or last part of the document that can lead to incorrect sentiment classification. To this end, we propose the using of an existing document representation method called paragraph vector as additional input features for Bi-LSTM. This vector provides information context of the document for each sequence processing. The paragraph vector is simply concatenated to each word vector of the document. This representation also helps to differentiate ambiguous Indonesian words. Bi-LSTM and paragraph vector were previously used as separate methods. Combining the two methods has shown a significant performance improvement of Indonesian sentiment analysis model. Several case studies on testing data showed that the proposed method can handle the sentiment phrases position problem encountered by Bi-LSTM.

## I. INTRODUCTION

Sentiment analysis as one of the topics in natural language processing has been greatly developed by the demand to know people opinion about a product, a service, or an issue in society. Research in this field is also supported by the development of technology such as social media and other online platforms that facilitate people to share their opinions and reviews. People may express their opinion towards an object or an event in a long text consist of one or more sentences. But then we want to know the conclusion of the sentiment expressed as a whole context. This is known as document level sentiment analysis.

Document level sentiment analysis was done with the assumption that each document describe a single entity [8]. In the scope of Indonesian language, several studies has been conducted to solve this problem by using conventional machine learning algorithms. Some of them use Support Vector Machine (SVM), Naïve Bayes Classifier, and Decision Tree [4, 11, 18]. Recently, deep learning has become a powerful technique to address this problem such as Bi-LSTM that showed a promising performance in sentiment classification task for non-Indonesian languages [1, 9, 10, 12]. However, only few studies in Indonesian sentiment analysis (document level) were conducted using deep learning technique and based on the research we have known so far, all of them are for aspect-based problem [3, 7]. Thus, we propose the using of Bi-LSTM to solve Indonesian sentiment analysis.

In Indonesian language, there are some ambiguous words with the same spelling but different meaning based on context. In order to differentiate those words, we also used a document representation called paragraph vector as input for Bi-LSTM. The paragraph vector was concatenated to each word vector of a document. This vector provides information context of the document for each sequence processed by Bi-LSTM. The resulting model can also overcome the problem in Bi-LSTM sequential behavior which tends to make sentiment predictions based on the first and/or last phrases of document and lost the document context.

This paper is organized in the following manner. Section 2 discusses related works. Section 3 discusses the proposed method. Section 4 shows the experimental results. Section 5 discusses the analysis and case study. Finally, the conclusion is in section 6.

## II. RELATED WORKS

Research for Indonesian sentiment analysis has been conducted for years. Wijaya et al. [11] compared three machine learning algorithm, Support Vector Machine, Naive Bayes, and Decision Tree for twitter sentiment analysis of Indonesian mobile operators. The experiment showed that SVM holds the highest accuracy rate of 83.33%. SVM also showed the best accuracy in experimental result of sentiment classification for Indonesian message in social media [6]. Lidya et al. [18] performed Indonesian sentiment analysis experiment using SVM and K-Nearest Neighbor (K-NN). Both SVM and K-NN received TF-IDF vector as input features. Fold cross validation showed that SVM also outperformed K-NN performance.

Farhan and Khodra [2] proposed Sentiment Specific Word Embedding (SSWE) for sentiment analysis on Indonesian review texts. Each text is represented by average word vector with TF-IDF. This method helps to reduce noise by words that have high frequency on text [20]. The model showed a promising performance but still inferior to SVM model with lexical features of TF-IDF. Lutfi et al. [4] also showed that SVM with TF-IDF produced a significant performance for Indonesian sentiment analysis with accuration rate of 93.65%.

TF-IDF is high dimensional sparse vector as an extension of Bag of Words (BoW) where the term frequency counts are discounted by the inverse document-frequencies [13].

However, TF-IDF does not consider words order. Two documents with the same composition of words can have the same TF-IDF vector representation, regardless of the difference in polarity. For example, sentence "*Saya suka karena tidak ada yang merokok*" ("I like it because nobody smokes") and sentence "*Saya tidak suka karena ada yang merokok*" ("I don't like it because someone smokes") have the same TF-IDF vector representation but the polarity is each other opposite. This will lead to incorrect classification.

In the other hand, SVM also encounter difficulty to predict sentiment of document with positive and negative sentiment expression (mixed polarity). SVM classifier find an optimal hyperplane as the solution to the learning problem [19]. The simplest formulation of SVM is linear function and to solve the non-linear problem the kernel tricks is used. Nevertheless, it is difficult to find hyperplane that perfectly separate positive and negative documents especially when it comes to large dataset. The model performance tends to remain constant in the increasing number of data. SVM also has low confidence level on predicting mixed polarity documents because those documents are located near the hyperplane. Thus, it is difficult for SVM to make conclusion about the sentiment of document.

Recently, many deep learning techniques have shown promising performance to solve sentiment analysis problem and can be applied for Indonesian language texts. Salinca [5] used Convolutional Neural Network (CNN) to solve sentiment classification on business reviews. The reviews are in English. In 2018, Bi-LSTM model has reach state-of-the-art for Bengali sentiment analysis, outperformed SVM and Decision Tree [1]. Bi-LSTM with word embedding as input features also showed significant performance on sentiment analysis compare to other several methods such as RNN, CNN, LSTM, Naive Bayes [9, 10, 12]. The method we proposed for building Indonesian sentiment analysis model is to use paragraph vector as additional input features of Bi-LSTM. Paragraph vector alone as a method to build sentiment analysis model has shown competitive performance with deep learning techniques in document classification [21, 22, 23].

## III. INDONESIAN SENTIMENT ANALYSIS

In our work, the sentiment analysis model were built using Bi-LSTM that widely known as one of deep learning techniques. It processes inputs as sequence of information. The inputs are documents consist of words. Each word is represented as low dimensional vector and processed sequentially by Bi-LSTM. The features extracted from documents are the hidden vectors produced after processing all words in documents. Due to Bi-LSTM sequential principle, the first and/or last phrases of document tend to give stronger influence on making sentiment predictions. To overcome this problem, an existing document representation method called paragraph vector was used as input features.

### A. Bi-LSTM

Bidirectional Long Short-Term Memory (Bi-LSTM) is used to learn the matrix representation of documents and create model for Indonesian sentiment analysis. LSTM itself was developed as an extension of standard recurrent neural network to overcome vanishing gradient problem [17]. LSTM architecture consist of three gates and a cell memory state that processed sequence of inputs in a directed cycle connection between neurons to store information. In Bi-LSTM, the sequence inputs are processed in two directions cycle, forward and backward timesteps.

The first Bi-LSTM model as in Fig. 1 was built using basic input features where each word is represented as word vector.

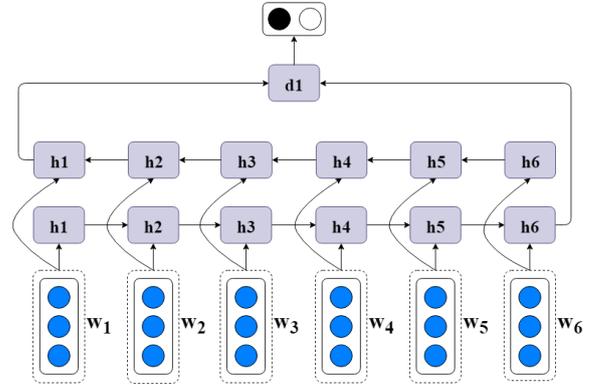

Fig. 1. The architecture of Bi-LSTM with word vectors {$w_1$, $w_2$, ..., $w_n$} from word embedding model. { $h_1$, $h_2$, ..., $h_n$ } are hidden vectors of Bi-LSTM.

The vector is low dimensional vector obtained from pre-trained Indonesian word embedding. The input sequences are computed sequentially by Bi-LSTM and produces a hidden vector of the last word in document. This vector is then connected to a sigmoid layer to do a binary sentiment classification. The resulting model performance will be compared to our proposed model to examine the effectiveness of the model.

### B. Bi-LSTM with Paragraph Vector

Paragraph vector was initially introduced by Le and Mikolov [14]. The vector is used as a memory to store information about the topic or context of document. The context is represented as a fixed-length vector. There are two approaches to learn the paragraph vector of a document. The first method is distributed memory model (PV-DM) as shown in Fig. 2.

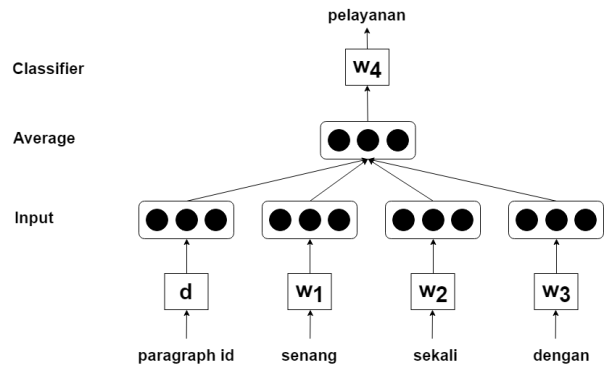

Fig. 2. Framework for learning paragraph vector through distributed memory method.

PV-DM learns the context representation of a document through processing window that was shifted throughout the document [14]. The model tries to predict the next word based on context (paragraph id) and known words. PV-DM was trained to learn the context and words vector by using stochastic gradient descent through backpropagation. The second method used to obtain paragraph vector of a document is distributed bag of words (PV-DBOW) as shown in Fig. 3.

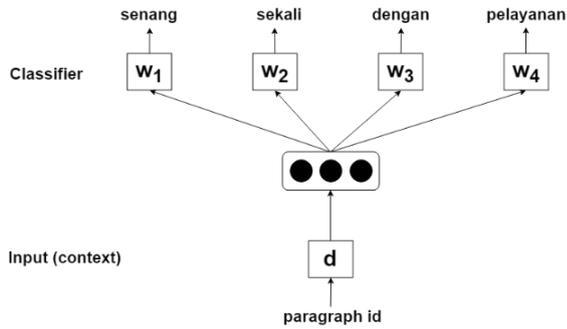

Fig. 3. Framework for learning paragraph vector through distributed bag of words method.

PV-DBOW learns the document vectors by predicting several consecutive words in a document given the paragraph id of document [14]. Processing window took out random samples from documents and used it as inputs to train PV-DBOW model.

We used Gensim Doc2Vec [15] library to build the matrix of paragraph vector. The resulting vectors were combined with word vectors of pre-trained Indonesian word embedding as input of Bi-LSTM. Fig. 4 illustrates that each word vector in a document is concatenated to the same paragraph vector that represent information context of the document. Our model used both PV-DM and PV-DBOW representation with each vector dimension 100. Thus, the paragraph vector length for a document is 200. Meanwhile, the length of each word vector is 500.

|  | Paragraph Vector | | | | Word Vector | | | |
|---|---|---|---|---|---|---|---|---|
| 1st word | 0.1 | -0.8 | 0.3 | ...... | 0.3 | -0.2 | 0.1 | ...... |
| 2nd word | 0.1 | -0.8 | 0.3 | ...... | 0.1 | 1.45 | 0.05 | ...... |
| 3rd word | 0.1 | -0.8 | 0.3 | ...... | 0.12 | -0.3 | 0.8 | ...... |
| 4th word | 0.1 | -0.8 | 0.3 | ...... | 1.16 | 0.08 | -0.6 | ...... |
| ...... | ...... | ...... | ...... | ...... | ...... | ...... | ...... | ...... |
| nth word | 0.1 | -0.8 | 0.3 | ...... | 2.05 | 1.35 | 1.2 | ...... |

Fig. 4. The illustration of a document representation using paragraph vector and word vector from word embedding model.

The words in a document are represented as 700 length vectors. Two same words in different documents will have different representation based on its context. Each word has information context of document from the paragraph vector. All words in the matrix representation of document then processed sequentially by Bi-LSTM.

An experiment with similar type of inputs for deep neural network had been conducted in aspect-based sentiment analysis [16]. The model used average word vector of aspect sentiment words as information context. Ruder et al. [16] combined this vector with each word vector by several methods such as summation, concatenation, and multiplication. It showed that concatenation yields the best result. Thus, we applied concatenation in our model to combine the paragraph vectors and word vectors as Bi-LSTM input features. The architecture of Bi-LSTM model that uses paragraph vectors and word vectors as input is shown in Fig. 5.

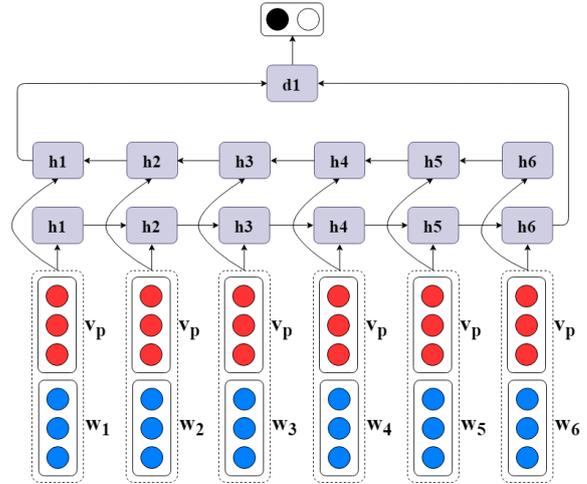

Fig. 5. The architecture of Bi-LSTM with paragraph vector ($v_p$) and word vectors {$w_1, w_2, ..., w_n$}. { $h_1, h_2, ..., h_n$ } are hidden vectors of Bi-LSTM.

IV. EXPERIMENT

In building Indonesian sentiment analysis model using Bi-LSTM and paragraph vector, we conducted experiment by doing parameter tuning on the hidden size of Bi-LSTM. The resulting model was then compared to the Bi-LSTM model without paragraph vector. We also compared our model with SVM for baseline as it showed significant performance in many previous Indonesian sentiment analysis experiments.

*A. Dataset*

The dataset we used to build each sentiment analysis model consist of separate training data and test data with polarity statistics as shown in Table I.

TABLE I. DATASET STATISTICS

| Dataset | Positive | Negative | Total |
|---|---|---|---|
| Training data | 7151 | 3830 | 10981 |
| Test data | 208 | 204 | 412 |

The dataset is a collection of texts or documents in Indonesian language obtained from Twitter, Zomato, TripAdvisor, Facebook, Instagram, Qraved. The dataset was crawled and then annotated by several Indonesian linguists. It consists of documents labeled positive and negative with statistics as shown in Table 1. The training data has vocabulary size of 15908 and document with the longest sequences consist of 92 word sequences. The test data has vocabulary size of 2194 and document with the longest sequences consist of 70 word sequences. Those documents discuss about various topics such as politic issues, social events, e-commerces, booking applications, foods, etc. Each document in the dataset has gone through language preprocessing such as emoticon deletion and word normalization.

*B. Experiment*

For experiment, the training data was divided into training data for configuration and validation data with a proportion of 90% and 10% respectively. In the learning process, in order to obtain the best model configuration for Bi-LSTM, we did early stopping to prevent overfitting. We also applied a 0.5 dropout rate, binary cross-entropy evaluation, and used Adam optimizer. The testing data is then used to compare the resulting model performance.

For the SVM baseline, we re-run the code of Farhan and Khodra [2] Indonesian sentiment analysis program with the same dataset. The program can be run with several features such as bag of words, TF-IDF, and average word vector. SVM with TF-IDF as features showed the best performance among the baseline methods. We then compared the performance of SVM with our model. From the experiment result in Table II, we can see that both Bi-LSTM models show a significant performance for Indonesian sentiment analysis.

TABLE II.  EVALUATION COMPARISON OF DIFFERENT MODELS

| Model | Precision | Recall | F1-Score |
| --- | --- | --- | --- |
| SVM (TF-IDF) | 0.7977 | 0.8878 | 0.8658 |
| Bi-LSTM (WE) | 0.9166 | 0.9126 | 0.9125 |
| **Bi-LSTM (PV+WE)** | **0.9384** | **0.9369** | **0.9369** |

They outperformed the SVM baseline because Bi-LSTM has the advantage of using word embedding to represent words that only occur in testing data. The use of paragraph vector as input features has improved Bi-LSTM performance.

## V. ANALYSIS AND CASE STUDY

We performed manual analysis to compare the performance of each sentiment analysis model. SVM makes a sentiment classification by searching the best hyperplane to separate positive and negative documents on dataset. It appears that SVM has difficulty on predicting mixed-polarity documents that have both positive and negative expressions/phrases in one document.

For example, the following document "*senang sekali gue dengan pelayanan komplain di di saat penjual tidak menggubris tokopedia memberikan solusi dengan cepat pasti nya dengan investigasi yang tidak berlarut larut. ya walau harus berakhir dengan pengembalian dana dana. jadi gue tidak pernah bosan buat belanja di tokopedia.*" ("I was very happy with the complaint service at the time the seller ignored then tokopedia provide a solution quickly and surely with an insoluble investigation. yes, although it must end with a refund of funds. so I never get bored for shopping in Tokopedia."), expresses positive and negative opinion about the whole e-commerce transaction. SVM could not make a correct conclusion about sentiment of the document while both Bi-LSTM model could predict the sentiment correctly.

The proposed model, Bi-LSTM (PV+WE) showed a significant performance improvement of Bi-LSTM by only adding paragraph vector as input features. Therefore, we examined several documents which its sentiment correctly predicted by Bi-LSTM (PV+WE) but incorrectly predicted by Bi-LSTM (WE) to investigate the advantage of the model we propose.

We found that several documents that incorrectly predicted by Bi-LSTM (WE) express their sentiment mainly in the beginning phrases or middle phrases of documents. As we learned from Bi-LSTM architecture, the first and/or last part of input sequence appears to have stronger features because it was processed in forward and backward sequential order. This causes Bi-LSTM tends to make predictions based on the first and last part of the document. Therefore, when the sentiment expression lies in the initial phrase or middle phrase of document, Bi-LSTM (WE) model may lose the context of the document and predict sentiment incorrectly. Concatenating paragraph vector to each word vector of documents provides information context of document. This helps Bi-LSTM to predict sentiments when the sentiment expressions is not disclosed at the end of the document.

To investigate this hypothesis, we took five sample documents from the testing data and made modifications. We modified the document by moving phrases that express sentiment from the beginning or middle part of document to the end of document without changing its meaning. One of the example is we modified document "*sudah lama tidak minum ultra milk rasa stroberi. pas sekarang minum merasa enak sekali tidak tahu kenapa. ketagihan. rasanya ingin beli sekardus gede buat diminum sendiri.*" ("long time not drinking strawberry flavored ultra milk. drink it right now feels so good I don't know why. addicted. it feels like I want to buy a carton of milk to drink alone.") into "*sudah lama tidak minum ultra milk rasa stroberi. rasanya ingin beli sekardus gede buat diminum sendiri. pas sekarang minum merasa enak sekali tidak tahu kenapa. ketagihan.*" ("long time not drinking strawberry flavored ultra milk. it feels like I want to buy a carton of milk to drink alone. drink it right now feels so good I don't know why. addicted.").

We performed sentiment predictions using Bi-LSTM (WE) and Bi-LSTM (PV+WE) for the modified documents. It turns out that Bi-LSTM (WE) which previously predicted the sentiment of the original document incorrectly, gave a correct prediction of the documents after modification. Meanwhile, Bi-LSTM (PV+WE) correctly predicts sentiments before and after modification. This shows that the use of paragraph vectors has improved Bi-LSTM performance by providing information context for each part of the document.

## VI. CONCLUSION

In this work, we propose Bi-LSTM with paragraph vectors and word vectors obtained from word embedding model to solve Indonesian sentiment analysis problem. The paragraph vectors are combined with each word vector in the document and form an input feature matrix. The experimental result shows that by simply adding paragraph vectors as Bi-LSTM input feature provides a significant performance improvement in sentiment classification. The proposed model also outperforms the SVM baseline which has become a state-of-the-art in many Indonesian sentiment analysis experiments.

Bi-LSTM processes input as a sequence of information and extracts features from documents word by word in sequence. Thus, the first and/or last phrases of document appear to have stronger features for sentiment predictions and cause Bi-LSTM to lose the information context of document. The paragraph vector in each word representation helps the model to maintain the information context and helps to distinguish ambiguos words in Indonesian language.

REFERENCES

[1] Abdullah Aziz Sharfuddin, Md. Nafis Tihami, and Md. Saiful Islam. "A Deep Recurrent Neural Network with BiLSTM model for Sentiment Classification," Proceedings of International Conference on Bangla Speech and Language Processing (ICBSLP), 2018.

[2] Ahmad Naufal Farhan and Masayu Leylia Khodra. "Sentiment-specific word embedding for Indonesian sentiment analysis," Proceedings of International Conference on Advanced Informatics: Concept Theory and Applications (ICAICTA), 2017.

[3] Alson Cahyadi and Masayu Leylia Khodra. "Aspect-Based Sentiment Analysis Using Convolutional Neural Network and Bidirectional Long Short-Term Memory," Proceedings of 5th International Conference on Advanced Informatics: Concept Theory and Applications (ICAICTA), 2018.

[4] Anang Anggono Lutfi, Adhistya Erna Permanasari, and Silmi Fauziati. "Sentiment Analysis in the Sales Review of Indonesian Marketplace by Utilizing Support Vector Machine," Journal of Information Systems Engineering and Business Intelligence, 2018.

[5] Andreea Salinca. "Convolutional Neural Networks for Sentiment Classification on Business Reviews," Computing Research Repository, arXiv: 1710.05978, version 1, 2017.

[6] Aqsath Rasyid Naradhipa and Ayu Purwarianti. "Sentiment classification for Indonesian message in social media," Proceedings of the International Conference on Cloud Computing and Social Networking (ICCCSN), 2012.

[7] Arfinda Ilmania, Abdurahman, Ayu Purwarianti, and Samuel Cahyawijaya. "Aspect Detection and Sentiment Classification using Deep Neural Network for Indonesian Aspect-Based Sentiment Analysis," Proceedings of International Conference on Asian Language Processing (IALP), 2018.

[8] Bing Liu. "Sentiment Analysis and Opinion Mining," University of Illinois, Chicago, 2012.

[9] Duyu Tang, Bing Qin, and Ting Liu. "Document Modeling with Gated Recurrent Neural Network for Sentiment Classification," Proceedings of the Conference on Empirical Methods in Natural Language Processing (EMNLP), 2015.

[10] Guixian Xu, Yueting Meng, Xiaoyu Qiu, Ziheng Yu, and Xu Wu. "Sentiment Analysis of Comment Texts Based on BiLSTM," Journal of Institute of Electrical and Electronics Engineers (IEEE) Access, vol. 7, pp. 51522–51532, 2019.

[11] Hansen Wijaya, Alva Erwin, Amin Soetomo, and Maulahikmah Galinium. "Twitter Sentiment Analysis and Insight for Indonesian Mobile Operators," Proceedings of the Information Systems International Conference (ISICO), 2013.

[12] Junhao Zhou, Yue Lu, Hong-Ning Dai, Hao Wang, and Hong Xiao. "Sentiment Analysis of Chinese Microblog Based on Stacked Bidirectional LSTM," Journal of Institute of Electrical and Electronics Engineers (IEEE) Access, vol. 7, pp. 38856–38866, 2019.

[13] Karen Sparck Jones. "A statistical interpretation of term specificity and its application in retrieval," Journal of documentation, vol. 28, issue: 1, pp.11-21, 1972.

[14] Quoc V. Le and Tomas Mikolov. "Distributed Representations of Sentences and Documents," Computing Research Repository, arXiv: 1405.4053, version 1, 2014.

[15] Radim Rehurek and Petr Sojka. "Software Framework for Topic Modelling with Large Corpora," Proceedings of the LREC Workshop on New Challenges for NLP Frameworks, 2010.

[16] Sebastian Ruder, Parsa Ghaffari, and John G. Breslin. INSIGHT-1 at SemEval-2016 Task 5: Deep Learning for Multilingual Aspect-based Sentiment Analysis, Computing Research Repository, arXiv: 1609.02748, version 1, 2016.

[17] Sepp Hochreiter and Jurgen Schmidhuber. "Long short-term memory," Journal of Neural computation, 9(8):1735–1780, 1997.

[18] Syahfitri Kartika Lidya, Opim Salim Sitompul, and Syahril Efendi. "Sentiment Analysis Pada Teks Bahasa Indonesia Menggunakan Support Vector Machine (SVM) dan K-Nearest Neighbor (K-NN)," Proceedings of Seminar Nasional Teknologi Informasi dan Komunikasi (SENTIKA), 2015.

[19] Theodoros Evgeniou and Massimiliano Pontil. "Support Vector Machines: Theory and Applications," Lecture Notes in Computer Science, vol. 2049, Springer, Berlin, Heidelberg, 2001.

[20] Tom Kenter, Alexey Borisov, and Maarten de Rijke. "Siamese CBOW: Optimizing Word Embeddings for Sentence Representations," Computing Research Repository, arXiv: 1606.04640, version 1, 2016.

[21] Yoon Kim. "Convolutional Neural Networks for Sentence Classification," Proceedings of Conference on Empirical Methods in Natural Language Processing (EMNLP), 2014.

[22] Yukun Zhu, Ryan Kiros, Ruslan Salakhutdinov, Richard S. Zemel, Antonio Torralba, Raquel Urtasun, Sanja Fidler. "Skip-Thought Vectors," Computing Research Repository, arXiv:1506.06726, version 1, 2015.

[23] Zichao Yang, Diyi Yang, Chris Dyer, Xiaodong He, Alex Smola1, Eduard Hovy. "Hierarchical Attention Networks for Document Classification," Proceedings of NAACL-HLT, pages 1480–1489, 2016.